\pdfoutput=1

\documentclass[11pt]{article}

\usepackage[]{acl}
\usepackage{bm}
\usepackage{times}
\usepackage{latexsym}
\usepackage[noorphans,vskip=0.6ex]{quoting}
\usepackage{multirow}

\usepackage[T1]{fontenc}

\usepackage[utf8]{inputenc}

\usepackage{microtype}

\usepackage{amssymb}
\usepackage{tabularx}
\usepackage{graphicx}
\usepackage{booktabs}

\title{Detecting Unassimilated Borrowings in Spanish:\\ An Annotated Corpus and Approaches to Modeling}

\author{Elena Álvarez-Mellado \\
    NLP \& IR group \\ School of Computer Science, UNED \\
  \texttt{elena.alvarez@lsi.uned.es} \\\And
  Constantine Lignos \\
  Michtom School of Computer Science  \\
  Brandeis University \\
  \texttt{lignos@brandeis.edu} \\}

\begin{document}
\maketitle
\begin{abstract}
This work presents a new resource for borrowing identification and analyzes the performance and errors of several models on this task. We introduce a new annotated corpus of Spanish newswire rich in unassimilated lexical borrowings---words from one language that are introduced into another without orthographic adaptation---and use it to evaluate how several sequence labeling models (CRF, BiLSTM-CRF, and Transformer-based models) perform. The corpus contains 370,000 tokens and is larger, more borrowing-dense, OOV-rich, and topic-varied than previous corpora available for this task. Our results show that a BiLSTM-CRF model fed with subword embeddings along with either Transformer-based embeddings pretrained on codeswitched data or a combination of contextualized word embeddings outperforms results obtained by a multilingual BERT-based model.
\end{abstract}

\section{Introduction and related work}
Lexical borrowing is the process of bringing words from one language into another \citep{haugen1950analysis}.
Borrowings are a common source of out-of-vocabulary (OOV) words, and the task of detecting borrowings has proven to be useful both for lexicographic purposes and for NLP downstream tasks such as parsing \citep{alex2008automatic}, text-to-speech synthesis \citep{leidig2014automatic}, and machine translation \citep{tsvetkov2016cross}.

Recent work has approached the problem of extracting lexical borrowings in European languages such as German \citep{alex-2008-comparing,garley-hockenmaier-2012-beefmoves,leidig2014automatic}, Italian \citep{furiassi2007retrieval}, French \citep{alex2008automatic,chesley2010lexical}, Finnish \citep{mansikkaniemi2012unsupervised}, Norwegian \citep{andersen2012semi,losnegaard2012data}, and Spanish \citep{serigos2017applying}, with a particular focus on English lexical borrowings (often called \textit{anglicisms}).

Computational approaches to mixed-language data have traditionally framed the task of identifying the language of a word as a tagging problem, where every word in the sequence receives a language tag \citep{lignos2013toward,molina-etal-2016-overview,solorio-etal-2014-overview}.
As lexical borrowings can be single (e.g. \textit{app}, \textit{online}, \textit{smartphone}) or multi-token (e.g. \textit{machine learning}), they are a natural fit for chunking-style approaches.
\citet{alvarez2020lazaro} introduced chunking-based models for borrowing detection in Spanish media which were later improved \citep{alvarez2020annotated}, producing an F1 score of 86.41.

However, both the dataset and modeling approach used by \citet{alvarez2020annotated} had significant limitations.
The dataset focused exclusively on a single source of news and consisted only of headlines.
The number and variety of borrowings were limited, and there was a significant overlap in borrowings between the training set and the test set, which prevented assessment of whether the modeling approach was actually capable of generalizing to previously unseen borrowings.
Additionally, the best results were obtained by a CRF model, and more sophisticated approaches were not explored.

The contributions of this paper are a new corpus of Spanish annotated with unassimilated lexical borrowings and a detailed analysis of the performance of several sequence-labeling models trained on this corpus.
The models include a CRF, Transformer-based models, and a BiLSTM-CRF with different word and subword embeddings (including contextualized embeddings, BPE embeddings, character embeddings, and embeddings pretrained on codeswitched data).
The corpus contains 370,000 tokens and is larger and more topic-varied than previous resources.
The test set was designed to be as difficult as possible; it covers sources and dates not seen in the training set, includes a high number of OOV words (92\% of the borrowings in the test set are OOV) and is very borrowing-dense (20 borrowings per 1,000 tokens).

The dataset we present is publicly available\footnote{\url{https://github.com/lirondos/coalas}} and has been released under a CC BY-NC-SA-4.0 license. This dataset was used for the ADoBo shared task on automatic detection of borrowings at IberLEF 2021 \citep{mellado2021overview}.

\begin{table}[tb]
\centering
\small

\begin{tabular}[t]{llll}
\toprule
Media & Topics & Set(s) \\
\midrule
ElDiario.es & General newspaper & Train, Dev. \\
El orden mundial & Politics & Test \\
Cuarto poder &  Politics & Test\\
El salto &  Politics & Test\\
La Marea &  Politics & Test\\
Píkara Magazine & Feminism & Test\\
El blog salmón & Economy & Test\\
Pop rosa & Gossip & Test\\
Vida extra & Videogames & Test\\
Espinof & Cinema \& TV & Test\\
Xataka & Technology & Test\\
Xataka Ciencia & Technology & Test\\
Xataka Android & Technology & Test\\
Genbeta & Technology & Test\\
Microsiervos & Technology & Test\\
Agencia Sinc & Science & Test\\
Diario del viajero & Travel & Test\\
Bebe y más & Parenthood & Test\\
Vitónica & Lifestyle \& sports & Test\\
Foro atletismo & Sports & Test\\
Motor pasión & Automobiles & Test\\
\bottomrule
\end{tabular}
\caption{Sources included in each dataset split (URLs provided in Appendix~\ref{sec:app_sources})}
\label{tab:media}
\end{table}

\section{Data collection and annotation}

\subsection{Contrasting lexical borrowing with codeswitching}

Linguistic borrowing can be defined as the transference of linguistic elements between two languages. 
Borrowing and codeswitching have been described as a continuum \citep{clyne2003dynamics}. 

Lexical borrowing involves the incorporation of single lexical units from one language into another language and is usually accompanied by morphological and phonological modification to conform with the patterns of the recipient language \citep{onysko2007anglicisms,poplack1988social}. 

On the other hand, codeswitches are by definition not integrated into a recipient language, unlike established loanwords \citep{poplack2012does}.
While codeswitches require a substantial level of fluency, comply with grammatical restrictions in both languages, and are produced by bilingual speakers in bilingual discourses, lexical borrowings are words used by monolingual individuals that eventually become lexicalized and assimilated as part of the recipient language lexicon until the knowledge of ``foreign'' disappears \citep{lipski2005code}.

\subsection{Data selection}

\begin{table}[tb]
\centering
\small
\begin{tabular}[t]{lrrrr}
\toprule
Set & Tokens & \texttt{ENG} & \texttt{OTHER}  & Unique \\
\midrule
Training & 231,126  & 1,493 & 28 & 380 \\
Development & 82,578 & 306 & 49 & 316 \\
Test & 58,997 & 1,239 & 46 & 987 \\
\midrule
Total & 372,701 & 3,038 & 123 & 1,683 \\
\bottomrule
\end{tabular}
\caption{Corpus splits with counts}
\label{tab:corpus}
\end{table}

Our dataset consists of Spanish newswire annotated for unassimilated lexical borrowings.
All of the sources used are European Spanish online publications (newspapers, blogs, and news sites) published in Spain and written in European Spanish.

Data was collected separately for the training, development, and test sets to ensure minimal overlap in borrowings, topics, and time periods. 
The training set consists of a collection of articles appearing between August and December 2020 in \textit{elDiario.es}, a progressive online newspaper based in Spain.
The development set contains sentences in articles from January 2021 from the same source.

The data in the test set consisted of annotated sentences extracted in February and March 2021 from a diverse collection of online Spanish media that covers specialized topics rich in lexical borrowings and usually not covered by \textit{elDiario.es}, such as sports, gossip or videogames (see Table~\ref{tab:media}).

To focus annotation efforts for the training set on articles likely to contain unassimilated borrowings, the articles to be annotated were selected by first using a baseline model and were then human-annotated.
To detect potential borrowings, the CRF model and data from \citet{alvarez2020lazaro} was used along with a dictionary look-up pipeline. 
Articles that contained more than 5 borrowing candidates were selected for annotation.

The main goal of data selection for the development and test sets was to create borrowing-dense, OOV-rich datasets, allowing for better assessment of generalization.
To that end, the annotation was based on sentences instead of full articles.
If a sentence contained a word either flagged as a borrowing by the CRF model, contained in a wordlist of English, or simply not present in the training set, it was selected for annotation. This data selection approach ensured a high number of borrowings and OOV words, both borrowings and non-borrowings.
While the training set contains 6 borrowings per 1,000 tokens, the test set contains 20 borrowings per 1,000 tokens.
Additionally, 90\% of the unique borrowings in the development set were OOV (not present in training).
92\% of the borrowings in the test set did not appear in training (see Table~\ref{tab:corpus}).

\subsection{Annotation process}
The corpus was annotated with BIO encoding using Doccano \citep{doccano} by a native speaker of Spanish with a background in linguistic annotation (see Appendix~\ref{sec:data_statement}).
The annotation guidelines (provided in Appendix~\ref{guidelines}) were based on those of \citet{alvarez2020annotated} but were expanded to account for a wider diversity of topics.  
Following \citeauthor{serigos2017applying}'s observations and \citeauthor{alvarez2020annotated}'s work, English lexical borrowings were labeled \texttt{ENG}, other borrowings were labeled \texttt{OTHER}.
Here is an example from the training set:\footnote{``Benching: being on your crush's bench while someone else plays in the starting lineup.''}

\begin{quoting}\small
\textbf{Benching} [\texttt{ENG}], estar en el banquillo de tu \textbf{crush} [\texttt{ENG}] mientras otro juega de titular.
\end{quoting}

In order to assess the quality of the guidelines and the annotation, a sample of 9,110 tokens from 450 sentences (60\% from the test set, 20\% from training, 20\% from development) was divided among a group of 9 linguists for double annotation.
The mean inter-annotation agreement computed by Cohen's kappa was 0.91, which is above the 0.8 threshold of reliable annotation \citep{artstein2008inter}.

\subsection{Limitations}

Like all resources, this resource has significant limitations that its users should be aware of.
The corpus consists exclusively of news published in Spain and written in European Spanish.
This fact by no means implies the assumption that European Spanish represents the whole of the Spanish language.

The notion of assimilation is usage-based and community-dependant, and thus the dataset we present and the annotation guidelines that were followed were designed to capture a very specific phenomena at a given time and in a given place: unassimilated borrowings in the Spanish press.

In order to establish whether a given word has been assimilated or not, the annotation guidelines rely on lexicographic sources such as the prescriptivist \textit{Diccionario de la Lengua Española} \cite{dle} by the Royal Spanish Academy, a dictionary that aims to cover world-wide Spanish but whose Spain-centric criteria has been previously pointed out \cite{blanch1995americanismo,fernandez2014lexicografia}.
In addition, prior work has suggested that Spanish from Spain may have a higher tendency of anglicism-usage than other Spanish dialects \cite{mcclelland2021brief}.
Consequently, we limit the scope of the dataset to European Spanish not because we consider that this variety represents the whole of the Spanish-speaking community, but because we consider that the approach we have taken here may not account adequately for the whole diversity in borrowing assimilation within the Spanish-speaking world.

\subsection{Licensing}

Our annotation is licensed with a permissive CC BY-NC-SA-4.0 license. 
With one exception, the sources included in our dataset release their content under Creative Commons licenses that allow for reusing and redistributing the material for non commercial purposes. 
This was a major point when deciding which news sites would be included in the dataset. 
The exception is the source \textit{Microsiervos}, whose content we use with explicit permission from the copyright holder. 
Our annotation is “stand-off” annotation that does not create a derivative work under Creative Commons licenses, so ND licenses are not a problem for our resource. 
Table~\ref{tab:media_url} in Appendix~\ref{sec:app_sources} lists the URL and license type for each source.

\section{Modeling}

\begin{table}[tb]
\centering
\small
\begin{tabular}[t]{lrrr}
\toprule
Set & Precision & Recall & F1 \\
\midrule
Development &  &  & \\
\hspace{0.5cm} \texttt{ALL} &  $ 74.13 $ & $ 59.72 $ & $ 66.15 $ \\
\hspace{0.5cm} \texttt{ENG} & $ 74.20 $ & $ 68.63 $ & $ 71.31 $ \\
\hspace{0.5cm} \texttt{OTHER} & $ 66.67 $ & $ 4.08 $ & $ 7.69 $ \\
\midrule
Test &  &  & \\
\hspace{0.5cm} \texttt{ALL} & $ 77.89 $ & $ 43.04 $ &  $ 55.44 $ \\
\hspace{0.5cm} \texttt{ENG} & $ 78.09 $ & $ 44.31 $ &  $ 56.54 $ \\
\hspace{0.5cm} \texttt{OTHER} &  $ 57.14 $ & $ 8.70 $ &  $ 15.09 $  \\

\bottomrule
\end{tabular}
\caption{CRF performance on the development and test sets (results from a single run)}
\label{tab:CRF}
\end{table}

\begin{table*}[tb]
\centering
\small
\begin{tabular}{l*{7}r}
\toprule
& \multicolumn{3}{c}{\textbf{Development}} &  & \multicolumn{3}{c}{\textbf{Test}} \\ 
\cmidrule(lr){2-4} \cmidrule(lr){6-8}
& Precision & Recall & F1 & & Precision & Recall & F1 \\
\midrule
BETO & & &  & & & \\
\hspace{0.25cm} \texttt{ALL} & $73.36 \pm \hphantom{0}3.6$ & $73.46	\pm 1.6$ & $73.35	\pm \hphantom{0}1.5$  &  & $86.76 \pm \hphantom{0}1.3$ &  $75.50 \pm 2.8$ &  $80.71 \pm 1.5$  \\
\hspace{0.25cm} \texttt{ENG} & $74.30	\pm \hphantom{0}1.8$  & $84.05 \pm 1.6$ & $78.81 \pm \hphantom{0}1.6$ &  &  $87.33 \pm \hphantom{0}2.9$ & $77.99 \pm 1.5$ & $82.36	\pm 1.5$  \\
\hspace{0.25cm} \texttt{OTHER} & $47.24 \pm	22.2$ & $7.34	\pm 3.7$  & $11.93 \pm	\hphantom{0}4.9$ & & $36.12 \pm	10.1$ & $8.48 \pm 3.5$  & $13.23 \pm 3.6$ \\

mBERT & & &  & & & \\
\hspace{0.25cm} \texttt{ALL} &  $\bm{79.96} \pm	\hphantom{0}1.9$ &    $\bm{73.86} \pm 2.7$ &  $\bm{76.76}$ $\pm$ $\hphantom{0}2.0$  &  &  $\bm{88.89} \pm \hphantom{0}1.0$ & $\bm{76.16} \pm 1.6$ & $\bm{82.02}	\pm 1.0$ \\
\hspace{0.25cm} \texttt{ENG} & $80.25	\pm \hphantom{0}2.6$ & $84.31 \pm	1.9$ & $82.21 \pm	\hphantom{0}1.9$ &  &  $89.25 \pm \hphantom{0}1.6$ &    $78.85 \pm 1.0$ &  $83.64 \pm 1.0$  \\
\hspace{0.25cm} \texttt{OTHER} &  $66.18 \pm 16.5$ & $8.6 \pm	6.8$ & $14.41 \pm 10.0$ & & $45.30	\pm 11.3$ & $7.61 \pm	1.5$ & $12.84	\pm 2.3$ \\

\bottomrule
\end{tabular}
\caption{Mean and standard deviation of scores on the development and test sets for BETO and mBERT}
\label{tab:transformers}
\end{table*}

The corpus was used to evaluate four types of models for borrowing extraction: (1) a CRF model, (2) two Transformer-based models, (3) a BiLSTM-CRF model with different types of unadapted embeddings (word, BPE, and character embeddings) and (4) a BiLSTM-CRF model with previously fine-tuned Transformer-based embeddings pretrained on codeswitched data.
By \emph{unadapted} embeddings, we mean embeddings that have not been fine-tuned for the task of anglicism detection or a related task (e.g. codeswitching).

Evaluation for all models required extracted spans to match the annotation exactly in span and type to be correct.
Evaluation was performed with SeqScore \citep{palen-michel-etal-2021-seqscore}, using \texttt{conlleval}-style repair for invalid label sequences.
All models were trained using an AMD 2990WX CPU and a single RTX 2080 Ti GPU.

\subsection{Conditional random field model}
\label{section:crf}
As baseline model, we evaluated a CRF model with handcrafted features from \citet{alvarez2020lazaro}. 
The model was built using \texttt{pycrfsuite} \citep{korobov2014python}, a Python wrapper for \texttt{crfsuite} \citep{CRFsuite} that implements CRF for labeling sequential data. The model also uses the \texttt{Token} and \texttt{Span} utilities from \texttt{spaCy} library \citep{honnibal2017spacy}. The following handcrafted binary features from \citet{alvarez2020lazaro} were used for the model:\\
-- Bias: active on all tokens to set per-class bias\\
-- Token: the string of the token\\
-- Uppercase: active if the token is all uppercase\\
-- Titlecase: active if only the first character of the token is capitalized\\
-- Character trigram: an active feature for every trigram contained in the token\\
-- Quotation: active if the token is any type of quotation mark (` ' " `` '' « »)\\
-- Suffix: last three characters of the token\\
-- POS tag: part-of-speech tag of the token provided by \texttt{spaCy} utilities\\
-- Word shape: shape representation of the token provided by \texttt{spaCy} utilities\\
-- Word embedding: provided by Spanish word2vec 300 dimensional embeddings by \citet{cardellinoSBWCE}, one feature per dimension \\
-- URL: active if the token could be validated as a URL according to \texttt{spaCy} utilities\\
-- Email: active if the token could be validated as an email address by \texttt{spaCy} utilities\\
-- Twitter: active if the token could be validated as a possible Twitter special token: \texttt{\#hashtag} or \texttt{@username}

A window of two tokens in each direction was used for feature extraction.
Optimization was performed using L-BFGS, with the following hyperparameter values chosen following the best results from \citet{alvarez2020lazaro} were set: c1 = $0.05$, c2 = $0.01$. 
As shown in Table~\ref{tab:CRF}, the CRF produced an overall F1 score of 66.15 on the development set (P: 74.13, R: 59.72) and an overall F1 of 55.44 (P:~77.89, R:~43.04) on the test set.
The CRF results on our dataset are far below the F1 of 86.41 reported by \citet{alvarez2020lazaro}, showing the impact that a topically-diverse, OOV-rich dataset can have, especially on test set recall.
These results demonstrate that we have created a more difficult task and motivate using more sophisticated models.

\subsection{Transformer-based models}\label{transformer}

We evaluated two Transformer-based models:

\par -- BETO base cased model: a monolingual BERT model trained for Spanish \citep{CaneteCFP2020}

\par -- mBERT: multilingual BERT, trained on Wikipedia in 104 languages \citep{devlin-etal-2019-bert}

Both models were run using the \texttt{Transformers} library by HuggingFace \citep{wolf-etal-2020-transformers}.
The same default hyperparameters were used for both models: 3 epochs, batch size 16, and maximum sequence length 256.
Except where otherwise specified, we report results for 10 runs that use different random seeds for initialization.
We perform statistical significance testing using the Wilcoxon rank-sum test.

As shown in Table~\ref{tab:transformers}, the mBERT model (F1:~82.02) performed better than BETO (F1:~80.71), and the difference was statistically significant ($p = 0.027$).
Both models performed better on the test set than on the development set, despite the difference in topics between them, suggesting good generalization.
This is a remarkable difference with the CRF results, where the CRF performed substantially worse on the test set than on the development set. The limited number of \texttt{OTHER} examples  explains the high deviations in the results for this label. 

\subsection{BiLSTM-CRF}
\label{bilstm}

We explored several possibilities for a BiLSTM-CRF model fed with different types of word and subword embeddings. The purpose was to assess whether the combination of different embeddings that encode different linguistic information could outperform the Transformer-based models in Section~\ref{transformer}. All of our BiLSTM-CRF models were built using \texttt{Flair} \citep{akbik2018coling} with default hyperparameters (hidden size = 256, learning rate = 0.1, mini batch size = 32, max number of epochs = 150) and embeddings provided by \texttt{Flair}. 

\subsubsection{Preliminary embedding experiments}

We first ran exploratory experiments on the development set with different types of embeddings using \texttt{Flair} tuning functionalities. We explored the following embeddings: Transformer embeddings (mBERT and BETO), fastText embeddings \citep{bojanowski-etal-2017-enriching}, one-hot embeddings, byte pair embeddings \citep{heinzerling-strube-2018-bpemb}, and character embeddings \citep{lample-etal-2016-neural}.

The best results were obtained by a combination of mBERT embeddings and character embeddings (F1: 74.00), followed by a combination of BETO embeddings and character embeddings (F1: 72.09).
These results show that using contextualized embeddings unsurprisingly outperforms non-contextualized embeddings for this task, and that subword representation is important for the task of extracting borrowings that have not been adapted orthographically. The finding regarding the importance of subwords is consistent with previous work; feature ablation experiments for borrowing detection have shown that character trigram features contributed the most to the results obtained by a CRF model \citep{alvarez2020lazaro}. 

The worst result (F1: 39.21) was produced by a model fed with one-hot vectors, and the second-worst result was produced by a model fed exclusively with character embeddings.
While it performed poorly (F1: 41.65), this fully unlexicalized model outperformed one-hot embeddings, reinforcing the importance of subword information for the task of unassimilated borrowing extraction.

\begin{table*}[htb]
\centering
\small
\begin{tabular}{lccrrr}
\toprule
Word embedding & BPE embedding & Char embedding & Precision  & Recall & F1 \\
\midrule
\texttt{mBERT} & - & -  & $ 82.27 $ & $ 69.30 $ & $ 75.23 $   \\
\texttt{mBERT} & - & \checkmark  & $ 79.45 $ & $ 72.96 $ &  $ 76.06 $   \\
\texttt{mBERT} & \texttt{multi} & \checkmark &  $ 81.37 $ & $ 73.80 $ & $ 77.40 $ \\
\texttt{mBERT} & \texttt{es, en} & - &  $ 83.07 $  & $ 74.65 $ &  $ 78.64 $\\
\texttt{mBERT} & \texttt{es, en} & \checkmark & $ 80.83 $ & $ 77.18 $ & $ 78.96 $ \\
\texttt{BETO, BERT} & - & \checkmark & $ 81.44 $ & $ 76.62 $ & $ 78.96 $  \\
\texttt{BETO, BERT} & - & - & $ 81.87 $ &   $ 76.34 $ & $ 79.01 $  \\
\texttt{BETO, BERT} & \texttt{es, en} & \checkmark & $ 85.94 $ & $ \bm{77.46} $ & $ 81.48 $ \\
\texttt{BETO, BERT} & \texttt{es, en} & - & $ \bm{86.98} $ & $ 77.18 $ & $ \bm{81.79} $ \\
\bottomrule
\end{tabular}
\caption{Scores (\texttt{ALL} labels) on the development set using a BiLSTM-CRF model with different combinations of word and subword embeddings  (results from a single random seed)}
\label{tab:embeddings}
\end{table*}

\subsubsection{Optimally combining embeddings}
In light of the preliminary embedding experiments and our earlier experiments with Transformer-based models, we fed our BiLSTM-CRF model with different combinations of contextualized word embeddings (including English BERT embeddings from \citeauthor{devlin-etal-2019-bert}), byte-pair embeddings and character embeddings. 
Table~\ref{tab:embeddings} shows development set results from different combinations of embeddings.
The best overall F1 on the development set was obtained by the combination of BETO embeddings, BERT embeddings and byte-pair embeddings. The model fed with BETO embeddings, BERT embeddings, byte-pair embeddings and character embeddings ranked second.

Several things stand out from the results in Table~\ref{tab:embeddings}.
The BETO+BERT embedding combination consistently works better than mBERT embeddings, and BPE embeddings contribute to better results.
Character embeddings, however, seem to produce little effect at first glance.
Given the same model, adding character embeddings produced little changes in F1 or even slightly hurt the results.
Although character embeddings seem to make little difference in overall F1, recall was consistently higher in models that included character embeddings, and in fact, the model with BETO+BERT embeddings, BPE embeddings and character embeddings produced the highest recall overall (77.46).
This is an interesting finding, as our results from Sections~\ref{section:crf} and~\ref{transformer} as well as prior work \citep{alvarez2020lazaro} identified recall as weak for borrowing detection models.

\begin{table*}[tb]
\centering
\small
\resizebox{\textwidth}{!}{\begin{tabular}{l*{7}r}
\toprule
\multirow{2}{*}{\textbf{Embeddings}} & \multicolumn{3}{c}{\textbf{Development}} &  & \multicolumn{3}{c}{\textbf{Test}} \\ 
\cmidrule(lr){2-4} \cmidrule(lr){6-8}
& Precision & Recall & F1 & & Precision & Recall & F1 \\
\midrule
BETO+BERT and BPE & & &  & & & \\
\hspace{0.25cm} \texttt{ALL} &  $\bm{85.84} \pm	\hphantom{0}1.2$ &	$77.07 \pm	0.8$ &	$\bm{81.21} \pm	0.5$ & &	$\bm{90.00} \pm	0.8$ &	$76.89 \pm	1.8$	& $82.92 \pm	1.1$\\
\hspace{0.25cm} \texttt{ENG} & $86.15 \pm	\hphantom{0}1.1$	& $88.00	\pm 0.6$	& $87.05 \pm	0.6$ 	& & $90.20 \pm	1.9$	& $79.36	\pm 1.1$ &	$84.42	\pm 1.1$  \\
\hspace{0.25cm} \texttt{OTHER} & $72.81 \pm	13.7$ &	$8.8	\pm 0.9$	& $15.60	\pm 1.6$ & &	$62.68	\pm 8.0$	& $10.43	\pm 0.9$ &	$17.83 \pm	1.3$ \\\\

BETO+BERT, BPE, and char & & &  & & & \\
\hspace{0.25cm} \texttt{ALL} & $84.29\pm	\hphantom{0}1.0$ &	$\bm{78.06}\pm	0.9$ 	& $81.05\pm	0.5$ & &	$89.71\pm	0.9$ &	$\bm{78.34}\pm	1.1$ &	$\bm{83.63}\pm	0.7$\\
\hspace{0.25cm} \texttt{ENG} & $84.54\pm	\hphantom{0}1.0$ &	$89.05\pm	0.5$ &	$86.73\pm	0.5$ & &	$89.90\pm	1.1$	& $80.88\pm	0.7$ &	$85.14\pm	0.7$  \\
\hspace{0.25cm} \texttt{OTHER} & $73.50\pm	11.4$ & 	$9.38\pm	3.0$ & 	$16.44\pm	4.8$ & & 	$61.14\pm	7.9$ & 	$9.78\pm	1.8$ & 	$16.81\pm	2.9$\\
\bottomrule
\end{tabular}}
\caption{Mean and standard deviation of scores on the development and test sets using a BiLSTM-CRF model with BETO and BERT embeddings, BPE embeddings, and optionally character embeddings}
\label{tab:best_bilstm}
\end{table*}

The two best-performing models from Table~\ref{tab:embeddings} (BETO+BERT embeddings, BPE embeddings and optionally character embeddings) were evaluated on the test set.
Table~\ref{tab:best_bilstm} gives results per type on the development and test sets for these two models.
For both models, results on the test set were better (F1: 82.92, F1: 83.63) than on the development set (F1: 81.21, F1: 81.05).
Although the best F1 score on the development set was obtained with no character embeddings, when run on the test set the model with character embeddings obtained the best score; however, these differences did not show to be statistically significant. Recall, on the other hand, was higher when run with character embeddings (R: 78.34) than when run without them (R: 76.89), and the difference was statistically significant ($p = 0.019$). This finding again corroborates the positive impact that character information can have in recall when dealing with previously unseen borrowings.

\begin{table*}[tb]
\centering
\small
\resizebox{\textwidth}{!}{\begin{tabular}{l*{7}r}
\toprule
\multirow{2}{*}{\textbf{Embeddings}} & \multicolumn{3}{c}{\textbf{Development}} &  & \multicolumn{3}{c}{\textbf{Test}} \\ 
\cmidrule(lr){2-4} \cmidrule(lr){6-8}
& Precision & Recall & F1 & & Precision & Recall & F1 \\
\midrule
Codeswitch & & &  & & & \\
\hspace{0.25cm} \texttt{ALL} & $80.21 \pm	\hphantom{0}1.7$	 & $74.42 \pm	1.5$	& $77.18 \pm	0.5$ & &	$90.05 \pm	\hphantom{0}1.0$	& $76.76	\pm 3.0$	& $82.83	\pm 1.6$ \\
\hspace{0.25cm} \texttt{ENG} & $80.19 \pm	\hphantom{0}1.6$ &	$85.59 \pm 0.5$ &	$82.78 \pm	0.5$ & &	$90.05 \pm	\hphantom{0}3.1$ &	$79.37 \pm	1.6$ &	$84.33 \pm	1.6$ \\
\hspace{0.25cm} \texttt{OTHER} & $85.83 \pm	19.2$ &	$4.70 \pm	2.3$ &	$8.78 \pm	4.2$ & &	$90.00 \pm	12.9$ &	$6.52	\pm 0.0$ &	$12.14 \pm	0.1$  \\

Codeswitch + char  & & &  & & & \\
\hspace{0.25cm} \texttt{ALL} & $81.02 \pm	\hphantom{0}2.5$ &	$74.56 \pm	1.5$ &	$77.62 \pm	0.9$ &&	$89.92 \pm	\hphantom{0}0.7$ &	$77.34 \pm	2.2$ &	$83.13 \pm	1.2$   \\
\hspace{0.25cm} \texttt{ENG} &  $81.00 \pm	\hphantom{0}1.6$ &	$85.91 \pm	0.9$ &	$83.34 \pm	0.9$ & &	$89.95 \pm	\hphantom{0}2.3$ &	$80.00	\pm 1.2$ &	$84.67 \pm	1.2$ \\
\hspace{0.25cm} \texttt{OTHER} & $73.00 \pm 41.6$ &	$3.67	\pm 2.7$ &	$6.91 \pm	4.9$ & &	$68.50 \pm	40.5$ &	$5.43 \pm	2.9$ &	$9.97	\pm 5.3$ \\

Codeswitch + BPE & & &  & & & \\
\hspace{0.25cm} \texttt{ALL} & $\bm{83.62} \pm	\hphantom{0}1.6$ &	$\bm{75.91} \pm	0.7$ &	$\bm{79.57} \pm	0.7$ & &	$90.43 \pm	\hphantom{0}0.7$ &	$78.55 \pm	1.3$ &	$84.06 \pm	0.7$\\
\hspace{0.25cm} \texttt{ENG} & $83.54 \pm	\hphantom{0}0.7$ &	$86.86 \pm	0.8$ &	$85.16 \pm	0.8$ & &	$90.57	\pm \hphantom{0}1.3$ &	$81.14 \pm	0.7$ &	$85.59 \pm	0.7$  \\
\hspace{0.25cm} \texttt{OTHER} &  $94.28 \pm	12.0$ &	$7.55 \pm	2.1$ &	$13.84 \pm	3.6$ & &	$67.17 \pm	\hphantom{0}8.3$ &	$8.70 \pm	1.4$ &	$15.30 \pm	2.2$ \\

Codeswitch + BPE + char   & & &  & & & \\
\hspace{0.25cm} \texttt{ALL} & $82.88 \pm	\hphantom{0}1.8$ &	$75.70 \pm	1.3$ &	$79.10 \pm	0.9$  & &	$\bm{90.60} \pm	\hphantom{0}0.7$ &	$\bm{78.72} \pm	2.5$ &	$\bm{84.22} \pm	1.6$   \\
\hspace{0.25cm} \texttt{ENG} & $82.90 \pm	\hphantom{0}1.4$ &	$86.57 \pm	1.0$ &	$84.66 \pm	1.0$ &	& $90.76 \pm	\hphantom{0}2.6$ &	$81.32 \pm	1.6$ &	$85.76 \pm	1.6$  \\
\hspace{0.25cm} \texttt{OTHER} & $87.23 \pm	14.5$ &	$7.75 \pm	2.8$ &	$14.03 \pm	4.7$ & &	$66.50 \pm	17.5$ &	$8.70 \pm	2.3$ &	$15.13 \pm	3.5$  \\
\bottomrule
\end{tabular}}
\caption{Mean and standard deviation of scores on the development and test sets for a BiLSTM-CRF model with combinations of codeswitch embeddings, BPE embeddings, and character embeddings}
\label{tab:codeswitch}
\end{table*}

\subsection{Transfer learning from codeswitching}

Finally, we decided to explore whether detecting unassimilated lexical borrowings could be framed as transfer learning from language identification in codeswitching.
As before, we ran a BiLSTM-CRF model using \texttt{Flair}, but instead of using the unadapted Transformer embeddings, we used codeswitch embeddings \cite{codeswich-embeddings}, fine-tuned Transformer-based embeddings pretrained for language identification on the Spanish-English section of the LinCE dataset \cite{aguilar-etal-2020-lince}.

Table~\ref{tab:codeswitch} gives results for these models. 
The two best-performing models were the BiLSTM-CRF with codeswitch and BPE embeddings (F1: 84.06) and the BiLSTM-CRF model with codeswitch, BPE and character embeddings (F1: 84.22). 
The differences between these two models did not show to be statistically significant, but the difference with the best-performing model with unadapted embeddings from Section \ref{bilstm} (F1: 83.63) was statistically significant ($p = 0.018$).
These  two best-performing models however obtained worse results on the development set than those obtained by the best-performing models from Section \ref{bilstm}.

Adding BPE embeddings showed to improve F1 score by around 1 point compared to either feeding the model with only codeswitch (F1: 82.83) or only codeswitch and character embeddings  (F1: 83.13), and the differences were statistically significant in both cases ($p = 0.024$, $p = 0.018$).

It should be noted  that this transfer learning approach is indirectly using more data than just the training data from our initial corpus, as the codeswitch-based BiLSTM-CRF models benefit from the labeled data seen during pretraining for the language-identification task.

\section{Error analysis}

We compared the different results produced by the best performing model of each type on the test set: (1) the mBERT model, (2) the BiLSTM-CRF with BERT+BETO, BPE and character embeddings and (3) the BiLSTM-CRF model with codeswitch, BPE and character embeddings.
We divide the error analysis into two sections.
We first analyze errors that were made by all three models, with the aim of discovering which instances of the dataset were challenging for all models.
We then analyze unique answers (both correct and incorrect) per model, with the aim of gaining insight on what are the unique characteristics of each model in comparison with other models.

\subsection{Errors made by all models}

\subsubsection{Borrowings labeled as \texttt{O}}

There were 137 tokens in the test set that were incorrectly labeled as \texttt{O} by all three models. 103 of these were of type \texttt{ENG}, 34 were of type \texttt{OTHER}. These errors can be classified as follows \\
\noindent -- Borrowings in upper case (12), which tend to be mistaken by models with proper nouns:
    


    \begin{quote}\small
Análisis de empresa  basados en \textbf{Big Data} [\texttt{ENG}].\footnote{``Business analytics based on Big Data''}
\end{quote}
\noindent -- Borrowings in sentence-initial position (9), which  were titlecased and therefore consistently mislabeled as \texttt{O}:

\begin{quote}\small
\textbf{Youtuber} [\texttt{ENG}], mujer y afroamericana: Candace Owen podría ser la alternativa a Trump.\footnote{``Youtuber, woman and African-American: Candace Owen could be the alternative to Trump''}
\end{quote}



\noindent Sentence-initial borrowings are particularly tricky, as models tend to confuse these with foreign named entities. In fact, prior work on anglicism detection based on dictionary lookup \citep{serigos2017applying} stated that borrowings in sentence-initial position were rare in Spanish and consequently chose to ignore all foreign words in sentence-initial position under the assumption that they could be considered named entities. However, these examples (and the difficulty they pose for models) prove that sentence-initial borrowings are not rare and therefore should not be overlooked.\\
-- Borrowings that also happen to be words in Spanish (8), such as the word \textit{primer}, that is a borrowing found in makeup articles (\textit{un primer hidratante}, ``a hydrating primer'') but also happens to be a fully Spanish adjective meaning ``first'' (\textit{primer premio}, ``first prize''). Borrowings like these are still treated as fully unassimilated borrowings by speakers, even when the form is exactly the same as an already-existing Spanish word and were a common source of mislabeling, especially partial mismatches in multitoken borrowings: \textit{red} (which exists in Spanish meaning ``net'') in \textit{red carpet}, \textit{tractor} in \textit{tractor pulling}  or \textit{total} in \textit{total look}. \\
-- Borrowings that could pass as Spanish words (58):
most of the misslabeled borrowings were words that do not exist in Spanish but that could orthographically pass for a Spanish word. That is the case of words like \textit{burpees} (hypothetically, a conjugated form of the non-existing verb \textit{burpear}),  \textit{gimbal}, \textit{mules}, \textit{bromance} or \textit{nude}.\\
-- Other borrowings (50): a high number of mislabeled borrowings were borrowings that were orthographically implausible in Spanish, such as \textit{trenchs}, \textit{multipads},  \textit{hypes}, \textit{riff}, \textit{scrunchie} or \textit{mint}. The fact that none of our models were  able to correctly classify these orthographically implausible examples leaves the door open to further exploration of character-based models and investigating character-level perplexity as a source of information.

\subsubsection{Non-borrowings labeled as borrowings}
29 tokens were incorrectly labeled as borrowings by all three models. These errors can be classified in the following groups:\\
    -- Metalinguistic usage and reported speech: a foreign word or sentence that appears in the text to refer to something someone said or wrote.
    \begin{quoting}\small
Escribir ``\textbf{icon pack}'' [\texttt{ENG}] en el buscador.\footnote{``Type `icon pack' on the search box''}
\end{quoting}

\noindent -- Lower-cased proper nouns: such as websites.
    \begin{quoting}\small
Acceder a la página \textbf{flywithkarolg} [\texttt{ENG}]\footnote{``Access the website flywithkarolg''}
\end{quoting}


\noindent -- Computer commands: the test set included blog posts about technology, which mentioned computer commands (such as \textit{sudo apt-get update}) that were consistently mistaken by our models as borrowings. These may seem like an extreme case---after all, computer commands do contain English words---but they are a good example of the real data that a borrowing-detection system may encounter.

\noindent -- Foreign words within proper nouns: lower-cased foreign words that were part of multitoken proper nouns.
    \begin{quoting}\small
    La serie ``10.000 
 \textbf{ships} [\texttt{ENG}]'' cuenta la odisea de la princesa Nymeria.\footnote{``The series `10,000 ships' tells the story of princess Nymeria''}
\end{quoting}

\noindent -- Acronyms and acronym expansions:
    \begin{quoting}\small
    El entrenamiento HITT
 (\textbf{high intensity interval training} [\texttt{ENG}])\footnote{``HITT training (High-intensity interval training)''}
\end{quoting}

\noindent -- Assimilated borrowings: certain borrowings that are already considered by RAE's dictionary as fully assimilated were labeled by all models as anglicisms.
    \begin{quoting}\small
    Labios rojos, a juego con el \textbf{top} [\texttt{ENG}].\footnote{``Red lips, matching top''}
\end{quoting}

\subsubsection{Type confusion}

Three tokens of type \texttt{OTHER} were marked by all models as \texttt{ENG}. There were no \texttt{ENG} borrowings that were labeled as \texttt{OTHER} by all three models. 

    \begin{quoting}\small
    Había  \textbf{buffet} [\texttt{ENG}] libre.\footnote{``There was a free buffet''}
\end{quoting}

\subsection{Unique answers per model}
We now summarize the unique mistakes and correct answers made per model, with the aim of understanding what data points were handled uniquely well or badly by each model.

\subsubsection{mBERT}

There were 46 tokens that were incorrectly labeled as borrowings only by the mBERT model.
These include foreign words used in reported speech or acronym expansion (21),  proper names (11) and already assimilated borrowings (7).

There  were 27 tokens that were correctly labeled only by the mBERT model.
The mBERT model was particularly good at detecting the full span of multitoken borrowings as in \textit{no knead bread}, \textit{total white}, \textit{wide leg} or \textit{kettlebell swings} (which were only partially detected by other models) and at detecting borrowings that could pass for Spanish words (such as \textit{fashionista}, \textit{samples}, \textit{vocoder}).
In addition, the mBERT model also correctly labeled as \texttt{O} 12 tokens that the other two models mistook as borrowings, including morphologically adapted anglicisms, such as \textit{craftear} (Spanish infinitive of the verb \textit{to craft}), \textit{crackear} (from \textit{to crack}) or \textit{lookazo} (augmentative of the noun \textit{look}).

\subsubsection{BiLSTM-CRF with unadapted embeddings}

There were 23 tokens that were incorrectly labeled as borrowings solely by this model, the most common types being assimilated borrowings (such as \textit{fan}, \textit{clon}) and Spanish words (\textit{fiestones}) (9 each).

32 tokens were correctly labeled as borrowings only by this model. These include borrowings that could pass for Spanish words (\textit{camel}, \textit{canvas}). In addition, this model also correctly labeled as \texttt{O} 6 tokens that the other two mistook as borrowings, including old borrowings that are considered today as fully assimilated (such as \textit{films} or \textit{sake}) or the usage of \textit{post} as a prefix of Latin origin (as in \textit{post-producción}), which other models mistook with the English word \textit{post}.

\begin{table*}[tb]
\centering
\small
\resizebox{\textwidth}{!}{\begin{tabular}{lccc*{7}r}
\toprule
\multirow{2}{*}{\textbf{Model}} &  \multirow{2}{*}{\textbf{Word emb}} & \multirow{2}{*}{\textbf{BPE emb}} & \multirow{2}{*}{\textbf{Char emb}} &  \multicolumn{3}{c}{\textbf{Development}} &  & \multicolumn{3}{c}{\textbf{Test}} \\ 
\cmidrule(lr){5-7} \cmidrule(lr){9-11}
& & & & Precision & Recall & F1 & & Precision & Recall & F1 \\
\midrule
CRF & \texttt{w2v (spa)} & - & - & $ 74.13 $ & $ 59.72 $ & $ 66.15 $ & & $ 77.89 $ & $ 43.04 $ & $ 55.44 $ \\
BETO & - & - & - & $ 73.36 $ &	$ 73.46 $ &	$ 73.35 $ & &	$ 86.76 $ &	$ 75.50 $ &	$ 80.71 $ \\
mBERT & - & - &  - & $ 79.96 $ &	$ 73.86 $ &	$ 76.76 $ & &	$ 88.89 $ &	$ 76.16 $	& $ 82.02 $ \\
BiLSTM-CRF & \texttt{BETO+BERT} & \texttt{en, es} & - & $ \bm{85.84} $ &	$ 77.07 $	& $ \bm{81.21} $ & &	$ 90.00 $ &	$ 76.89 $ &	$ 82.92 $ \\
BiLSTM-CRF & \texttt{BETO+BERT} & \texttt{en, es} & \checkmark & $ 84.29 $ &	$ \bm{78.06} $ &	$ 81.05 $ & &	$ 89.71 $	& $ 78.34 $ &	$ 83.63 $ \\
BiLSTM-CRF & \texttt{Codeswitch} & - & - & $ 80.21 $ &	$ 74.42 $ &	$ 77.18 $ & &	$ 90.05 $  &	$ 76.76 $ &	$ 82.83 $  \\
BiLSTM-CRF & \texttt{Codeswitch} & - & \checkmark & $ 81.02 $	& $ 74.56 $ &	$ 77.62 $ & &	$ 89.92 $ &	$ 77.34 $ &	$ 83.13 $\\
BiLSTM-CRF & \texttt{Codeswitch} & \texttt{en, es} & - & $ 83.62 $ &	$ 75.91 $ &	$ 79.57 $ & &	$ 90.43 $ &	$ 78.55 $ &	$ 84.06 $  \\
BiLSTM-CRF & \texttt{Codeswitch} & \texttt{en, es} & \checkmark & $ 82.88 $ &	$ 75.70 $	& $ 79.10 $	&  & $ \bm{90.60} $	& $ \bm{78.72} $ &	$ \bm{84.22} $\\
\bottomrule
\end{tabular}}
\caption{Scores (\texttt{ALL} labels) for the development and test sets across all models. For the CRF, results from a single run are reported. For all other models, the score reported is the mean calculated over 10 runs with different random seeds (see Tables \ref{tab:transformers}, \ref{tab:best_bilstm}, and \ref{tab:codeswitch} for standard deviations).}
\label{tab:results_summary}
\end{table*}

\subsubsection{BiLSTM-CRF with codeswitch embeddings}

The codeswitch-based system incorrectly labeled 18 tokens as borrowings, including  proper nouns (7), such as \textit{Baby Spice}, and  fully asimilated borrowings (5), such as \textit{jersey}, \textit{relax} or \textit{tutorial}.

This model correctly labeled 27 tokens that were mistakenly ignored by other models, including multitoken borrowings (\textit{dark and gritty}, \textit{red carpet}) and other borrowings that were non-compliant with Spanish orthographic rules but that were however ignored by other models (\textit{messy}, \textit{athleisure}, \textit{multitouch}, \textit{workaholic}).

The codeswitch-based model also correctly labeled as \texttt{O} 16 tokens that the other two models labeled as borrowings, including acronym expansions, lower-cased proper names and orthographically unorthodox Spanish words, such as the ideophone \textit{tiki-taka} or \textit{shavales} (a non-standard writing form of the word \textit{chavales}, ``guys'').

\section{Discussion}

Table~\ref{tab:results_summary} provides a summary of our results.
As we have seen, the diversity of topics and the presence of OOV words in the dataset can have a remarkable impact on results. 
The CRF model---which in previous work had reported an F1 score of 86---saw its performance drop to a 55 when dealing with our  dataset, despite the fact that both datasets consisted of journalistic European Spanish texts. 

On the other hand, neural models (Transformer-based and BiLSTM-CRF) performed better. 
All of them performed better on the test set than on the development set, which shows good generalization ability.
The BiLSTM-CRF model fed with different combinations of Transformer-based word embeddings and subword embeddings outperformed multilingual BERT and Spanish monolingual BETO. The model fed with codeswitch, BPE, and character embeddings ranked first and was significantly better than the result obtained by the model fed with BETO+BERT, BPE, and character embeddings.


Our error analysis shows that recall was a weak point for all models we examined. 
Among false negatives, upper-case borrowings (such as \textit{Big Data}) and borrowings in sentence-initial position (in titlecase) were frequent, as they tend to be mistaken with named entities. 
This finding suggests that borrowings with capitalized initial should not be overlooked. 
Similarly, words that exist both in English and Spanish (like \textit{primer} or \textit{red}) are not rare and were also a common source of error. 
Adding character embeddings produced a statistically significant improvement in recall, which opens the door to future work. 
 
Concurrently with the work presented on this paper, \citet{rosa2021futility} explored using supplementary training on intermediate labeled-data tasks (such as POS, NER, codeswitching and language identification) along with multilingual Transformer-based models to the task of detecting borrowings. Alternatively, \citet{jiang2021bert4ever} used data augmentation to train a CRF model for the same task.
\section{Conclusion}

We have introduced a new corpus of Spanish newswire annotated with unassimilated lexical borrowings. The test set has a high number of OOV borrowings---92\% of unique borrowings in the test set were not seen during training---and is more borrowing-dense and varied than resources previously available. We have used the dataset to explore several sequence labeling models (CRF, BiLSTM-CRF, and Transformer-based models) for the task of extracting lexical borrowings in a high-OOV setting. 
Results show that a BiLSTM-CRF model fed with Transformer-based embeddings pretrained on codeswitched data along subword embeddings produced the best results (F1: 84.22, 84.06), followed by a combination of contextualized word embeddings and subword embeddings (F1: 83.63). These models outperformed prior models for this task (CRF F1: 55.44) and multilingual Transformer-based models (mBERT F1: 82.02).

\section{Ethical considerations}
In this paper we have introduced an annotated dataset and models for detecting unassimilated borrowings in Spanish. The dataset is openly-licensed, and detailed annotation guidelines are provided (Appendix~\ref{guidelines}).
Appendix~\ref{sec:data_statement} includes a data statement that provides information regarding the curation rationale, annotator demographics, text characteristics, etc. of the dataset we have presented.
We hope these resources will contribute to bringing more attention to borrowing extraction, a task that has been little explored in the field of NLP but that can be of great help to lexicographers and linguists studying language change.

However, the resources we have presented should not be considered a full depiction of either the process of borrowing or the Spanish language in general.
We have identified four important considerations that any future systems that build off this research should be aware of.

    \paragraph{The process of borrowing.} Borrowing is a complex  phenomenon that can manifest at all linguistic levels (phonological, morphological, lexical, syntactic, semantic, pragmatic). 
    This work is exclusively concerned with lexical borrowings.
    Furthermore, in this work we have taken a synchronic approach to borrowing: we deal with borrowings that are considered as such in a given dialect and at a given point in time. 
    The process of borrowing assimilation is a diachronic process, and the notion of what is perceived as unassimilated can vary across time and varieties. 
    As a result, our dataset and models may not be suitable to account for partially assimilated borrowings or even for unassimilated borrowings in a different time period.  
    \paragraph{Language variety.} The dataset we have presented is exclusively composed of European Spanish journalistic texts. In addition, the guidelines we have described  were designed to capture a very specific phenomena: unassimilated borrowings in the Spanish press.
    In fact, the annotation guidelines rely on sources such as \textit{Diccionario de la Lengua Española}, a lexicographic source whose Spain-centric criteria has been previously pointed out \cite{blanch1995americanismo,fernandez2014lexicografia}.
    Consequently, the scope of our work is restricted to unassimilated borrowings in journalistic European Spanish. Our dataset and models may not translate adequately to other Spanish-speaking areas or genres.

\paragraph{The preeminence of written language.} In our work, the notion of what a borrowing is is heavily influenced by how a word is written. 
According to our guidelines, a word like \textit{meeting} will be considered unassimilated, while the Spanish form \textit{mitin} will be considered assimilated. 
These preferences in writing may indirectly reveal how well-established a loanword is or how foreign it is perceived by the speaker. 
But it is questionable that these two forms necessarily represent a difference in pronunciation or linguistic status in the speaker's mental lexicon. 
How a word is written can be helpful for the purpose of detecting novel anglicisms in written text, but ideally one would not establish a definition of borrowing solely based on lexicographic, corpus-derived or orthotypographic cues. These are all valuable pieces of information, but they only represent an indirect evidence of the status that the word holds in the lexicon. 
After all, speakers will identify a word as an anglicism (and use it as such), regardless of whether the word is written in a text or used in speech. 

On the other hand, the lexicographic fact that a word came from another language may not be enough as a criterion to establish the notion of borrowing.
Speakers use words all the time without necessarily knowing where they came from or how long ago they were incorporated into the language. 
The origin of the word may just be a piece of trivia that is totally irrelevant or unknown to the speaker at the time of speaking, so the etymological origin of the word might not be enough to account for the difference among borrowings. 
In fact, what lies at the core
of the unassimilated versus assimilated distinction is the awareness of speakers when they use a certain word \cite{poplack1988social}. 
The notion of what a borrowing is lies within the brain of the speaker, and 
in this work we are only indirectly observing that status through written form. Therefore our definition of borrowing and assimilation cannot be regarded as perfect or universal. 

\paragraph{Ideas about linguistic purity.} The purpose of this project is to analyze the usage of borrowings in the Spanish press. This project does not seek to promote or stigmatise the usage of borrowings, or those who use them.
The motivation behind our research is not to defend an alleged linguistic purity, but to study the phenomenon of lexical borrowing from a descriptive and data-driven point of view.

\section*{Acknowledgments}
The authors would like to thank Carlota de Benito Moreno, Jorge Diz Pico, Nacho Esteban Fernández, Gloria Gil, 
Clara Manrique, Rocío Morón González, Aarón Pérez Bernabeu, Monserrat Rius and Miguel Sánchez Ibáñez for their assessment of the annotation quality.

\bibliography{anthology,custom}
\bibliographystyle{acl_natbib}
\vfill\null
\pagebreak

\appendix

\section{Data sources}
\label{sec:app_sources}
See Table~\ref{tab:media_url} for the URLs and licenses of the sources used in the dataset.

\begin{table*}[tb]
\centering
\small

\begin{tabular}[t]{lll}
\toprule
Media & URL & License \\
\midrule
elDiario.es & \url{https://www.eldiario.es/} & CC BY-NC 4.0\\
Agencia Sinc & \url{https://www.agenciasinc.es/} & CC BY 4.0\\
El Salto & \url{https://www.elsaltodiario.com/} & CC BY-SA 3.0\\
La Marea & \url{https://www.lamarea.com/} & CC BY-SA 3.0\\
Cuarto poder & \url{https://www.cuartopoder.es/} & CC BY-NC 3.0\\
Genbeta & \url{https://www.genbeta.com/} & CC BY-NC 3.0\\

Bebe y más & \url{https://www.bebesymas.com/} & CC BY-NC 3.0\\
Diario del viajero & \url{https://www.diariodelviajero.com/} & CC BY-NC 3.0 \\
El blog salmón & \url{https://www.elblogsalmon.com/} & CC BY-NC 3.0 \\
Espinof & \url{https://www.espinof.com/} & CC BY-NC 3.0\\
Motor pasión & \url{https://www.motorpasion.com/} & CC BY-NC 3.0\\
Pop rosa & \url{https://www.poprosa.com/} & CC BY-NC 3.0 \\
Vida extra & \url{https://www.vidaextra.com/} & CC BY-NC 3.0\\
Vitónica & \url{https://www.vitonica.com/} & CC BY-NC 3.0\\
Xataka & \url{https://www.xataka.com/} & CC BY-NC 3.0\\
Xataka Ciencia & \url{https://www.xatakaciencia.com/} & CC BY-NC 3.0 \\
Xataka Android & \url{https://www.xatakandroid.com/} & CC BY-NC 3.0\\
El orden mundial & \url{https://elordenmundial.com/} & CC BY-NC-ND \\
Foro atletismo & \url{https://www.foroatletismo.com/} & CC BY-NC 2.1\\
Píkara Magazine & \url{https://www.pikaramagazine.com/} & CC BY-NC-ND \\
Microsiervos & \url{https://www.microsiervos.com/} & Used with written permission \\


\bottomrule
\end{tabular}
\caption{Media included in the corpus}
\label{tab:media_url}
\end{table*}

\section{Annotation guidelines}
\label{guidelines}

\subsection{Objective}
This document proposes a set of guidelines for annotating emergent unassimilated lexical borrowings, with a focus on English lexical borrowings (or anglicisms). 
The purpose of these annotation guidelines is to assist annotators to annotate unassimilated lexical borrowings from English that appear in Spanish newswire, i.e. words from English origin that are introduced into Spanish without any morphological or orthographic adaptation. 

This project approaches the phenomenon of lexical borrowing from a synchronic point of view, which means that we will not be annotating all words that have been borrowed at some point of the history of the Spanish language (like arabisms), but only those that have been recently imported and have not been integrated into the recipient language (in this case, Spanish).


\subsection{Tagset}
We will consider two possible tags for our annotation: \texttt{ENG}, for borrowings that come from the English language (or anglicisms), and \texttt{OTHER} for other borrowings that comply with the following guidelines but that come from languages other than English. 

\subsection{Defining an unassimilated lexical borrowing}
In this section we provide an overview of what words will be considered as unassimilated lexical borrowings for the sake of our annotation project.  
\subsubsection{Definition and scope}
The concept of \textit{linguistic borrowing} covers a wide range of linguistic phenomena. We will first provide a general overview of what lexical borrowing is and what will be understood as an anglicism within the scope of this project.

Lexical borrowing is the incorporation of single lexical units from one language (the donor language) into another language (the recipient language) and is usually accompanied by morphological and phonological modification to conform with the patterns of the recipient language \citep{haugen1950analysis,onysko2007anglicisms,poplack1988social}.

Anglicisms are lexical borrowings that come from the English language \citep{gomez1997towards,pratt1980anglicismo,gonzalez1999anglicisms,nunez2017up}. For our annotation project, we will focus on direct, unassimilated, emerging anglicisms, i.e. lexical borrowings from the English language into Spanish that have recently been imported and that have still not been assimilated into Spanish, that is, words like \textit{smartphone},  \textit{influencer}, \textit{hype}, \textit{lawfare} or \textit{reality show}.

Although this project focuses on lexical borrowings from English, we will also consider borrowings from other languages that comply with these guidelines. Borrowings from the English language will be annotated with the tag \texttt{ENG}, while borrowings from other languages shall be annotated with the tag \texttt{OTHER}:
\medbreak
\texttt{
… financiados a través de la plataforma de [crowdfunding](ENG) del club [gourmet](OTHER) que tengas más cerca}\footnote{Examples in these guidelines will display the lexical borrowing that should be labeled between square brackets, with the the corresponding tag in parentheses. Examples with no words marked with brackets will illustrate cases where no lexical borrowing should be tagged.}
\medbreak
Other types of borrowings, such as semantic calques, syntactic anglicisms or literal translations will be considered beyond the scope of these annotation project and will not be covered in these guidelines. 

\subsubsection{Types of lexical borrowing}
Lexical borrowings can be adapted (the spelling of the word is modified to comply with the phonological and orthographic patterns of the recipient language, as in \textit{fútbol} or \textit{tuit}) or unadapted (the word preserves its original spelling: \textit{millennial},  \textit{newsletter}, \textit{like}). For this annotation project, we will be focusing on unassimilated lexical borrowings: this means that adapted borrowings will be ignored and only unadapted borrowings will be tagged (see Section ~\ref{adapted} for a full description on the differences between adapted and unadapted borrowings). 

\subsubsection{Multiword borrowings}
Lexical borrowings can be both single-token units (\textit{online}, \textit{impeachment}), as well as multiword expressions (\textit{reality show}, \textit{best seller}). Multitoken borrowings will be labeled as one entity.
\medbreak

\texttt{imagina ser un ‘[tech bro]’ con millones de dólares (ENG)}

\medbreak

The annotation should however distinguish between a multitoken borrowing and adjacent borrowings. A phrase like \textit{signature look} is a multiword borrowing (the full phrase has been borrowed as a single unit) and should be annotated as such.
\medbreak

\texttt{para recrear su [total look] (ENG)}

\medbreak

However, a phrase like \textit{look sporty} follows the NAdj order that is typical of Spanish grammar (but impossible in English): these are in fact two separate borrowings (\textit{look} and \textit{sporty}) that have been borrowed independently and happen to be colocated in a phrase. The annotation should capture these nuances: 
\medbreak

\texttt{un [look] (ENG) [sporty] (ENG) perfecto}

\medbreak

\subsubsection{Origin of the borrowings}
Establishing the origin of a certain borrowings can sometimes be tricky, as the language of origin can sometimes be disputed. Additionally, certain borrowings might have originated in a certain language, but may have reached the recipient language through another language. 

In order to establish the origin of borrowings, the origin attributed by reference dictionaries and institutions \citep{dle,fundeu} will be followed. 

This means that words like \textit{junior} and \textit{senior} (whose frequency and perhaps even their pronunciation may have changed due to the influence of English) will still be considered as latinisms, as DLE registers their adaptated versions (\textit{júnior} and \textit{sénior}) as such (and mentions no English influence). Similarly, the word \textit{barista} might have entered the Spanish language via English, but RAE's Observatorio de Palabras considers it of Italian origin (and should therefore be annotated with OTHER label).

\begin{figure}[!b]
\begin{center}
\includegraphics[width=\columnwidth,keepaspectratio]{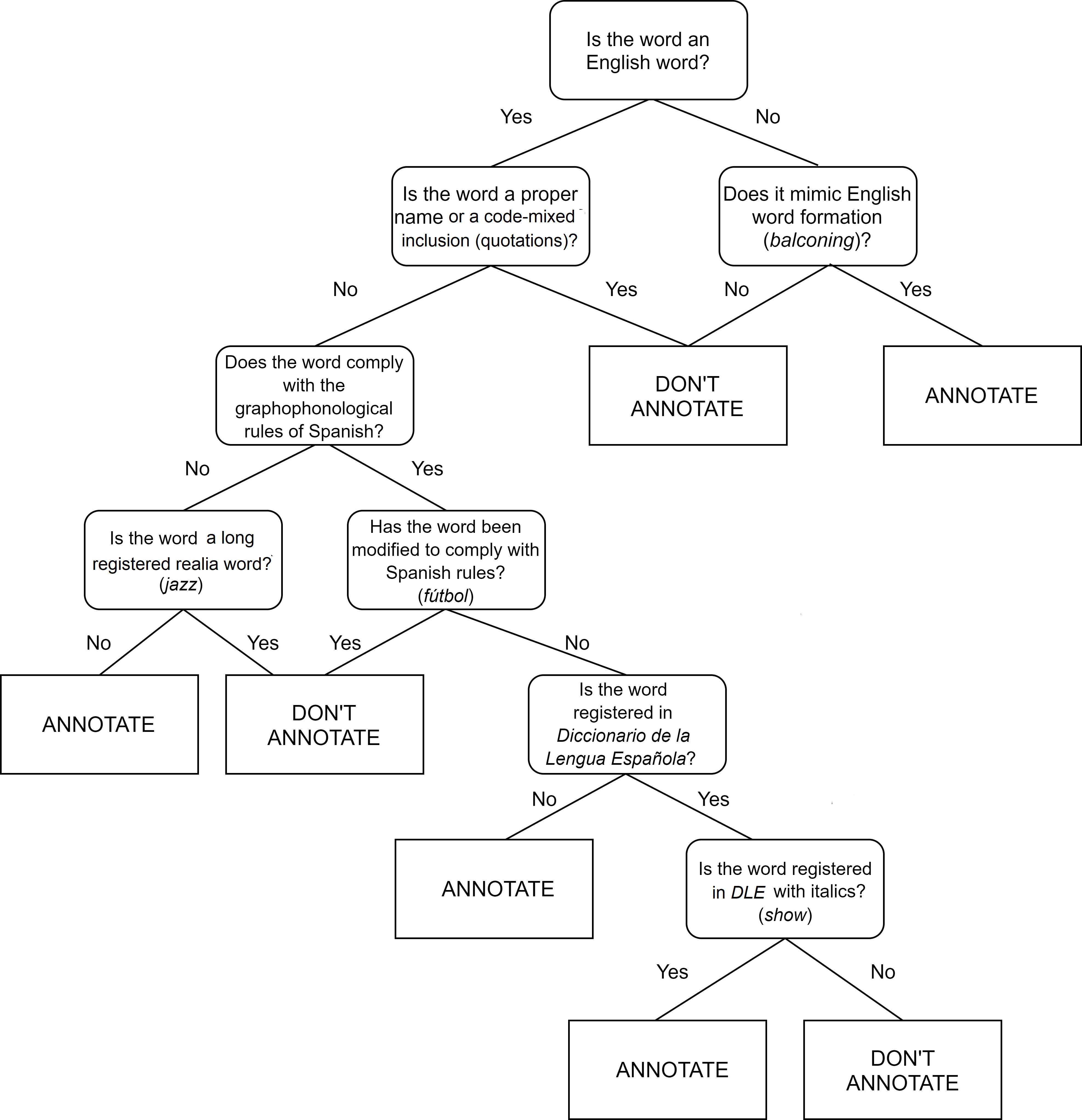}
\caption{Decision steps to follow during the annotation process to decide whether to annotate a word as an anglicism}
\label{fig:annotation}
\end{center}
\end{figure}

\subsection{What is not an unassimilated lexical borrowing}
In the previous section we provided an overview of what words will be considered as an unassimilated lexical borrowing for the sake of our annotation project. In this section we will cover what an unassimilated lexical borrowing is \textit{not}. 

There are several phenomena that are close enough to unassimilated borrowing and that can sometimes be mistaken with. In this section we will list what phenomena will not be considered as unassimilated lexical borrowings (and are therefore beyond the scope of our annotation project), as well as provide guidelines in order to distinguish these cases and adjudicate them. 

We will focus on three main phenomena: assimilated borrowings, proper names and code-mixed inclusions. 

Figure~\ref{fig:annotation} summarizes the decision steps that can be followed when deciding if a certain word should be labeled or not as a lexical borrowing. 

\subsubsection{Assimilated vs unassimilated borrowings}
This annotation project aims to capture unassimilated lexical borrowings. As a general rule, all unadapted lexical borrowings should be tagged. This means that direct borrowings that have not gone through any morphological or orthographic modification process should be labeled. 

Lexical adaptation, however, is a diachronic process and, as a result, what constitutes an unadapted borrowing is not clear-cut. The following guidelines define what borrowings will be considered as unassimilated (and therefore should be tagged) versus those that have already been integrated into the recipient language (and therefore should not be tagged).

\subsubsection{Adapted borrowings}\label{adapted}
Words that have already gone through orthographical or morphological adaptation (such as \textit{fútbol}, \textit{líder}, \textit{tuit} or \textit{espóiler}) will be considered assimilated and therefore should not be labeled. Partial adaptations (such as \textit{márketing}, where an accent has been added) will also be excluded.

 Borrowings that have not been adapted but whose original spelling complies with grapho-phonological rules of Spanish (and are therefore unlikely to be further adapted, such as \textit{bar}, \textit{fan}, \textit{web}, \textit{internet}, \textit{club}, \textit{set} or \textit{videoclip}) will be tagged as a borrowing or not not depending on how recent or emergent they are. In order to determine which unadapted, graphophonologically acceptable borrowings are to be annotated, the latest online version of the \textit{Diccionario de la lengua española} \citep{dle} will be consulted (as of February 2021)\footnote{\url{https://dle.rae.es/}}. If the DLE dictionary already registers the word with that meaning and with no italics or quotation marks, then it will be considered assimilated and therefore should not be tagged. 
 
 This means that a word like \textit{set} (when used to refer to a collection of things, a television studio or a part of a tennis match) will be considered assimilated because it is already registered in DLE dictionary with no italics, and therefore should not be labeled as \texttt{ENG}. On the other hand, a word like \textit{nude}, although its spelling also complies with Spanish graphophonological rules, will be considered an unassimilated borrowing because it has not been registered yet in the dictionary, and should therefore be tagged as such.
 
 \medbreak
\texttt{ganó el primer set}
\medbreak

\medbreak
\texttt{los tonos `[nude]' (ENG)}
\medbreak

It should be noted that this guideline only applies to lexical borrowings that comply with graphophonological rules of Spanish. Unadapted lexical borrowings that do not comply with graphophonological rules of Spanish (such as \textit{show}, \textit{look}, etc) will be tagged as borrowing, regardless of whether the word is included in the dictionary or not (although see section \ref{realia} for exceptions to this).

It is important to emphasize that, in order for an unadapted graphophonologically-compliant borrowing to be considered assimilated it should be registered in the dictionary both without italics and with the corresponding meaning. For instance, a word like \textit{top} (that is graphophonologically acceptable in Spanish) is registered in DLE with no italics, but it is only registered with the meaning of a piece of clothing. The word \textit{top} as referring to the upper part of something (as in \textit{top 5}) is not registered. Consequently, the borrowing \textit{top} will be considered assimilated when referring to the piece of clothing, but unassimilated when used to talk about the best elements of a ranking or the upper part of something.

\medbreak
\texttt{un top estampado}
\medbreak

\medbreak
\texttt{el [top] cinco de artistas (ENG)}
\medbreak

\medbreak
\texttt{la [top] desfiló (ENG)}
\medbreak

Similarly, the word \textit{post} will not be considered a borrowing when used as a prefix of Latin origin, but will be labeled with \texttt{ENG} when used to refer to something that is published on a social media platform.

\medbreak
\texttt{el mundo post pandemia}
\medbreak

\medbreak
\texttt{un [post] de Facebook (ENG)}
\medbreak

Additionally, assimilated borrowings can still be part of new unassimilated borrowings, in which case they will be labeled as such:  
\medbreak
\texttt{un [boys club] (ENG)}
\medbreak

\subsubsection{Words derived from foreign lexemes}
Words derived from foreign lexemes that do not comply with Spanish orthotactics but that have been morphologically derived following the Spanish paradigm (such as \textit{hacktivista}, \textit{randomizar}, \textit{shakespeariano}) will be considered assimilated and should therefore not be labeled as a borrowing.

Compound names where one of the lexemes is a borrowing will be labeled as a borrowing or not according to the degree of independence among the lexemes. A verb+noun compound (as \textit{caza-clicks}) will not be labeled as a borrowing, because the elements are not independent from one another. However, noun-noun compounds where each of the lexemes work can work independent from one can be labeled as borrowings: 

\medbreak
\texttt{una casa-[loft] (ENG)}
\medbreak

Similarly, prefixed borrowings will be labeled as a borrowing, as long as the borrowing keeps independence from the prefix:

\medbreak
\texttt{la ex [influencer] (ENG)}
\medbreak

For prefixed borrowings, it should be checked whether the prefix can also be considered part of the borrowing:

\medbreak
\texttt{los [nano influencers] (ENG)}
\medbreak

\subsubsection{Number inflection}
Unassimilated borrowings may be incorporated as invariable in number \textit{los master}, with the same plural inflection that they had in the donor language (\textit{los pappardelle}) or may form a new plural that is non-existant in the donor language  (\textit{los pappardelles}). For number inflection, we follow the same criteria that DLE \citep{dle} follows: a non-Italian plural like \textit{pizzas} is still regarded as unadapted (and therefore should be written italicized even when the true Italian plural would be \textit{pizze}). Consequently, non assimilated borrowings that have a non-cannonical plural inflection form will still be considered as an unassimilated borrowing and labeled as such.

\medbreak
\texttt{una serie de animación de [mechas] (OTHER)}
\medbreak

\subsubsection{Pseudoanglicisms}
Words that do not exist in English (or exist with a different meaning) but were coined following English morphological paradigm to imitate English words (such as \textit{footing} or \textit{balconing}) will be annotated as anglicisms.

\medbreak
\texttt{la imagen del `[balconing]' y las excursiones etílicas (ENG)}
\medbreak

\medbreak
\texttt{practicaba [footing] por la calle (ENG)}
\medbreak

\subsubsection{Realia words}\label{realia}
Borrowings that refer to culture-specific elements (often called \textit{realia words}) that were imported long ago but that have remained unadapted will not be tagged as borrowing. This means that if a borrowing is not adapted (i.e. its form remained exactly as it came from the donor language) but refers to a particular cultural object that came via the original language, that has been registered for a while in Spanish dictionaries and is not perceived as new anymore, then it will not be tagged as a borrowing, even if does not comply with graphophonologic rules of Spanish. 

The purpose of this guideline is to account for cultural terms such as \textit{pizza}, \textit{whisky}, \textit{jazz}, \textit{blues}, \textit{banjo} or \textit{sheriff}. These are all borrowings that are reluctant to be adapted or translated, even when they have been around in the Spanish language for long. The reason is that they refer to cultural inventions (the name was imported along with the object it refers to), and, given their cultural significance, they never competed with a Spanish equivalent and are seen as assimilated. 

Therefore, unadapted borrowings that refer to cultural innovations (such as music, cooking, sport names etc) and that have been registered for long in the Spanish language\footnote{RAE dictionary \url{https://dle.rae.es/}, Mapa de diccionarios \url{https://webfrl.rae.es/ntllet/SrvltGUILoginNtlletPub} and CREA \url{http://corpus.rae.es/creanet.html} and CORPES \url{https://webfrl.rae.es/CORPES/view/inicioExterno.view} can be consulted} will not be tagged as emergent borrowings.

It should be noted that this only applies to borrowings that have been around enough time to be registered in dictionaries. A word like \textit{hip hop} is a realia word, but it is still recent enough and has not been registered in the dictionary. In that case, it should be considered as unassimilated and tagged as such.  

\subsubsection{Latinisms}
Borrowings that were introduced directly from Latin language (such as \textit{deficit}, \textit{curriculum}, etc) will not be considered emergent and therefore will not be tagged as a borrowing. However, it should be noted that unassimilated borrowings from other languages that happen to have a Latin etymology and and that are introduced with a distinct meaning (such as \textit{adlib} or \textit{premium} etc) will still be tagged as borrowings.

\subsection{Borrowings vs names}

\subsubsection{Proper nouns}
Non-Spanish proper nouns will not be tagged as borrowings. These include:
\begin{itemize}
    \item person names: \textit{Bernie Sanders}.
\item organization names: \textit{WikiLeaks}. 
\item product names: \textit{Slack}. 
\item location names: \textit{Times Square}.
\item dates and celebrations: \textit{St. Patrick’s Day}, \textit{Black Friday}.
\item event names: \textit{Brexit}, \textit{procés}. 
\item social and political movements: \textit{Black Lives Matter}, \textit{MeToo}. 
\item treaties and documents: \textit{New Deal}, \textit{Privacy Shield}, \textit{French Tech Visa}. \item titles of cultural productions: \textit{Stranger Things}.
\end{itemize}

\subsubsection{Borrowings in proper nouns}
Borrowings that appear as part of proper nouns or named entities (such as book titles or organization names, as in \textit{Los Hermanos Podcast}) will not be labeled as borrowings.

\subsubsection{Proper nouns in borrowings}
Multiword borrowings and expressions can sometimes include proper nouns. Even when a proper noun in isolation cannot be considered a borrowing, proper nouns within a borrowed expression will be considered part of the borrowing, as long as the proper noun is part of the borrowing and is used following the grammar rules of the donor language (for example, in an English noun noun compound):  

\medbreak
\texttt{
Tecnología [made in Spain] (ENG)}
\medbreak

\medbreak
\texttt{
[Google cooking] (ENG)}
\medbreak

\subsubsection{Names of institutions and political roles}
Non-Spanish names that refer to political institutions (such as \textit{Parlament} or \textit{Bundestag}) or to political roles and figures (\textit{lehendakari}, \textit{president}, \textit{conseller}) will be excluded and will not be tagged as borrowings.

\subsubsection{Words derived from proper nouns}
Words derived from proper nouns (via metonymy or eponymy) will not be tagged as a borrowing, as long as the relation with the proper noun they come from is transparent to the speaker such as:
\begin{itemize}
    \item products: \textit{un iPhone}, \textit{un whatsapp}, \textit{un bizum}, \textit{un Scalextric}, \textit{el Satisfyer}.
    \item works of arts: \textit{un monet}
    \item characters: \textit{un frankestein}.
    
\end{itemize}

However, borrowings that originated from a proper noun in the donor language but entered the Spanish language as common nouns and are currently recognized as such, will be labeled as borrowings. In order to adjudicate which of these words are still used in Spanish as proper names and which are common nouns, dictionaries and other reference works can be consulted.

\subsubsection{Names of peoples or languages}
Names of peoples or languages (such as \textit{inuit}) will not be labeled as borrowings, even if the word is borrowed from another language and is not registered in Spanish dictionaries. 

\subsubsection{Ficticious creatures}
Unadapted names of fictitious creatures (such as \textit{hobbit} or \textit{troll}) will be labeled as a borrowing.
\medbreak
\texttt{
En un agujero en el suelo vivía un [hobbit] (ENG)}
\medbreak

\subsubsection{Scientific units}
Unadapted borrowings that refer to widespread scientific units (such as \textit{hertz}, \textit{newton}, \textit{byte}, etc) will be considered assimilated and should not be tagged as a borrowing

\subsubsection{Names of species}
Scientific names of a species (such as Latin names) will not be tagged as a lexical borrowing (\textit{anisakis}). Names of fruit, vegetable and plant varieties (such as \textit{manzana golden}, \textit{patatas Kennebec} or \textit{aguacate Hass}) will also be excluded.

\subsection{Borrowings vs other code-mixed inclusions}
Borrowing (using units from one language in another language) and codeswitching (intertwining segments of different languages in the same discourse) have frequently been described as a continuum \citep{clyne2003dynamics}, with a fuzzy frontier between the two. As a result, it can be difficult to tell the difference between borrowing and other code-mixed inclusions. The following guidelines can assist annotators adjudicate edge cases. 

When in doubt while dealing with code-mixed inclusion, the annotator may find it helpful to ask the following question as a rule of thumb: would it make sense to have this non-Spanish word registered in a dictionary of Spanish? If the answer is no (for instance, because the word reflects the literal quotation of what someone said or because the inclusions is metalinguistic usage rather than borrowing), then we are probably not in front of a borrowing but of another type of code-mixed inclusion (and should not be tagged as a borrowing).

\subsubsection{Acronyms and acronym expansions}
We consider acronyms to be a different phenomenon from borrowings. Consequently, acronyms will not be tagged as a borrowing, even if the acronym is of non-Spanish origin

\medbreak
\texttt{
un lector de CD}
\medbreak

An acronym however may be tagged as a borrowing if it appears as part of a borrowed multiword expression, as in \textit{CD player}, \textit{peak TV}, \textit{PC gaming}:
\medbreak
\texttt{
un [CD player] (ENG)}
\medbreak

Acronym expansions, that is, the expansion of an acronym into the words that form the acronym (that is usually added in between brackets after an acronym has been introduced) will also not be considered a borrowing:
\medbreak
\texttt{
La técnica de PCR (protein chain reaction)}
\medbreak

It is important to note that for a sequence to be considered as an acronym expansion it must appear after the acronym has been introduced and serve as a gloss to it (so that it expands what the letters in the acronym stand for). Usages where the full sequence is introduced in the text and later on acronymized for the sake of brevity can still be considered as borrowings. 
\medbreak

\texttt{
Utilizaron técnicas de [Machine Learning] (también conocido como ML) (ENG)} 
\medbreak

\subsubsection{Digits}
Similarly to proper nouns, digits in isolation cannot be considered borrowings. As a result, we cannot take for granted that digits within the surroundings of a borrowing will automatically be part of the borrowing. 

\medbreak
\texttt{
[top ten] (ENG)}
\medbreak

\texttt{
[top] 10 (ENG)}
\medbreak

However, if the word order of the tokens makes it clear that the digit is part of a multitoken borrowing (because the order complies with the grammatical structure of an English noun-noun compound), we can label it as part of the borrowing: 
\medbreak

\texttt{
los [10\% banks] (ENG)}
\medbreak

\subsubsection{Metalinguistic usage}
Non-Spanish words that appear to refer to the word itself in linguistic discourse and do not cover a lexical gap will not be tagged as a borrowing: \medbreak
\texttt{
El término viene de la palabra ``ghost'', que en inglés es `fantasma'}
\medbreak

It should be noted that the newer, less adapted, less transparent a new word is, the more likely that the speaker will be aware of the decoding difficulty it may pose to the reader and will decide to add some sort of metalinguistic strategy or awareness around it, in the form of metacomments, word-pointers, meaning explanations, etc (\textit{known as}, \textit{so called}). Borrowings with these types of signals will still be considered borrowings, as long as they are covering a lexical gap. 

True metalinguistic usage where the foreign word covers no lexical gap but exclusively provides linguistic information (such as etymological information) will not be considered a borrowing.

\subsubsection{Literal quotations}
Words or sequences in languages other than Spanish that are reflecting literally what someone said or wrote (as in a quotation, a statement or a slogan) will not be considered a borrowing. 
\medbreak
\texttt{
El eslogan `Make America Great Again'}
\medbreak

\medbreak
\texttt{
Es uno de los primeros resultados de Google cuando alguien busca "remote work in Spain" (trabajo en remoto en España).}
\medbreak

\subsubsection{Expressions}
In general terms, multiword borrowings will be tagged as borrowings. However, phrases and expressions that are not integrated into the sentence will be excluded. This means that autonomous expressions that are rather code switched sentences (rather than real borrowings) that work as a unit totally independently of the rest of the linguistic context (and that we would not expect to be registered in a dictionary) will not be considered or tagged as a borrowing.  

\medbreak
\texttt{
La innovación y la competencia tan escasas en la radiotelevisión o peor aún en Internet ("the winners takes all" o "most").}

\medbreak

\subsection{Limitations of these guidelines}
These guidelines are intended to assist annotators when labeling lexical borrowings. These guidelines, however, were created with a specific goal in mind (to capture unassimilated English lexical borrowings from a corpus of Spanish newswire) and may not be suitable if applied to a project with a different scope. These are some of the shortcomings and limitations that these guidelines may have.

\subsubsection{Text genre}
These guidelines were designed to specifically capture borrowings in a corpus of Spanish newswire. Newswire is a very specific genre of text that by no means represent the whole of a language \citep{plank2016non}. 

\subsubsection{Donor language}
These guidelines were created with English lexical borrowings in mind, which are the most frequent source of borrowing today in the Spanish press. Although the criteria can be applied to other languages as well (and in fact the annotation tagset we propose includes the tag \texttt{OTHER} to account for borrowings from other languages other than English), a more fine-grained approach would require further guidelines. 

\subsubsection{Synchronic approach to borrowing}
This project approaches emergent, unassimilated lexical borrowing in a synchronic fashion. The process of borrowing and the notion of assimilation is, however, time-dependent. A diachronic approach to lexical borrowing would require a wider scope, a different theoretical framework and an expanded set of criteria.

\subsubsection{Geographic variety}
The guidelines in this document were designed to capture borrowings used in Spanish newspapers, that is, written in the variety of Spanish that is spoken in Spain and may not be suitable to account for other dialects. 
For instance, according to the guidelines we have just introduced, a word like \textit{living} (that is used heavily in some Latin American varieties to refer to the living room) would be considered unassimilated. It is arguable whether these criteria would be suitable for a project that tried to capture emergent lexical borrowings in Argentinian text, for example.

\vfill\null

\section{Data statement}
\label{sec:data_statement}
We document the information concerning our dataset following the data statement format proposed by \citet{bender-friedman-2018-data}.\\ 
\textbf{Data set name:} Corpus of Anglicisms in the Spanish Press (COALAS) \\
\textbf{Data set developer:} Elena Álvarez-Mellado \\ 
\textbf{Dataset license:} Creative Commons Attribution-NonCommercial-ShareAlike 4.0 International (\href{https://creativecommons.org/licenses/by-nc-sa/4.0/}{CC BY-
NC-SA 4.0}) 
\\ 
\textbf{Link to  dataset:} \url{https://github.com/lirondos/coalas}

\subsection{Curation rationale}
The corpus consist of a collection of Spanish newspaper sentences. These sentences are annotated with unassimilated lexical borrowings. 

Data was collected separately for the training, development, and test sets to ensure minimal overlap in borrowings, topics, and time periods (see Section~\ref{sec:speech}). 

To focus annotation efforts for the training set on articles likely to contain unassimilated borrowings, the articles to be annotated were selected by first using a baseline model and were then human-annotated.
To detect potential borrowings, the CRF model and data from \citet{alvarez2020lazaro} was used along with a dictionary look-up pipeline. 
Articles that contained more than 5 borrowing candidates were selected for annotation.

The main goal of data selection for the development and test sets was to create borrowing-dense, OOV-rich datasets, allowing for better assessment of generalization.
To that end, the annotation was based on sentences instead of full articles.
If a sentence contained a word either flagged as a borrowing by the CRF model, contained in a wordlist of English, or simply not present in the training set, it was selected for annotation. This data selection approach ensured a high number of borrowings and OOV words, both borrowings and non-borrowings.
While the training set contains 6 borrowings per 1,000 tokens, the test set contains 20 borrowings per 1,000 tokens.
Additionally, 90\% of the unique borrowings in the development set were OOV (not present in training).
92\% of the borrowings in the test set did not appear in training (see Table~\ref{tab:corpus}).

\subsection{Language variety}
The language of this corpus is Standard Spanish (ISO 639-1 es). Since all the sources in the dataset are Spanish newspapers, the dialect is standard European Spanish. 

\subsection{Speaker demographic}
No detailed information was collected regarding the demographics of the authors of the collected sentences. However, we can infer that the authors of the text were Spanish journalists aged between 20-65.

\subsection{Annotator demographic}
The annotator was a 30-35 year-old female graduate student from Spain, who was trained in linguistics and computational linguistics, whose native language is Spanish and who had extensive professional proficiency in English.  

\subsection{Speech situation}
\label{sec:speech}
The training set consists of a collection of articles appearing between August and December 2020 in \textit{elDiario.es}, a progressive online newspaper based in Spain. 
The development set contains sentences in articles from January 2021 from the same source.

The test set consisted of annotated sentences extracted in February and March 2021 from a diverse collection of online Spanish media that covers specialized topics rich in lexical borrowings and usually not covered by elDiario.es, such as sports, gossip or videogames (see Table~\ref{tab:corpus}).

All the data came from written sources and was presumably edited according to the style guides of each source. 

\subsection{Text characteristics}
All the sentences in this dataset come from journalistic texts. The data in the training set and development set come from a general online newspaper. The data in the test set come from a diverse collection of online media covering specialized topics, which include politics, feminism, economy, gossip, videogames, cinema \& TV, technology, science, travel, parenthood, lifestyle, sports and automobiles.   
\subsection{Recording quality}
N/A
\subsection{Other}
N/A
\subsection{Provenance appendix}
See Table \ref{tab:media_url}.
\end{document}